\documentclass[11pt,letterpaper]{article}
\usepackage{emnlp2016}
\usepackage{times}
\usepackage{latexsym}
\usepackage{basecommon}
\usepackage{graphicx}
\usepackage{url}
\usepackage[font={small}]{caption}

\newcommand{\RNN}{\mathrm{\mathbf{RNN}}}

\newcommand{\BRNN}{\mathrm{\mathbf{BRNN}}}

\newcommand{\pfx}[1]{\ensuremath{w_{1:{#1}}}}
\newcommand{\goldpfx}[1]{\ensuremath{y_{1:{#1}}}}
\newcommand{\beampred}[2]{\ensuremath{\hat{y}_{1:{#1}}^{({#2})}}}
\DeclareMathOperator{\suk}{succ}
\DeclareMathOperator{\topK}{topK}

\newcommand{\nicein}{\ensuremath{\,{\in}\,}}
\newcommand{\niceq}{\ensuremath{\,{=}\,}}

\emnlpfinalcopy

\def\emnlppaperid{***}

\title{Sequence-to-Sequence Learning \\ as Beam-Search Optimization}

\author{Sam Wiseman \and Alexander M. Rush\\
 School of Engineering and Applied Sciences \\ Harvard University \\ Cambridge, MA, USA \\ {\tt \{swiseman,srush\}@seas.harvard.edu }}

\date{}

\begin{document}

\maketitle

\begin{abstract}
  Sequence-to-Sequence (seq2seq) modeling has rapidly become an
  important general-purpose NLP tool that has proven effective for
  many text-generation and sequence-labeling tasks. Seq2seq builds on deep neural language modeling and inherits its
  remarkable accuracy in estimating local, next-word
  distributions. In this work, we introduce a model and beam-search training
  scheme, based on the work of \newcite{daume05learning}, that extends
  seq2seq to learn global sequence scores. This
  structured approach avoids classical biases associated with local
  training and unifies the training loss with the test-time usage,
  while preserving the proven model architecture of seq2seq and
  its efficient training approach. We show that our system outperforms a
  highly-optimized attention-based seq2seq system and other baselines
  on three different sequence to sequence tasks: word ordering,
  parsing, and machine translation.
\end{abstract}

\section{Introduction}
Sequence-to-Sequence learning with deep neural networks (herein,
seq2seq)~\cite{sutskever11generating,sutskever14sequence} has rapidly
become a very useful and surprisingly general-purpose tool for natural
language processing. In addition to demonstrating impressive results
for machine translation \cite{bahdanau15neural}, roughly the same model and training
have also proven to be useful for sentence compression~\cite{filippova15sentence}, parsing \cite{vinyals15grammar}, and dialogue systems~\cite{serban16building}, and they additionally underlie other text generation applications, such as image or video captioning~\cite{vid2text,showattend}.

The dominant approach to training a seq2seq system is as
a conditional language model, with training maximizing the likelihood of each successive target word conditioned on the input sequence and
the \textit{gold} history of target words. Thus, training uses a strictly word-level loss, usually cross-entropy over the target vocabulary. This approach has proven to be very effective and efficient for training neural language models, and seq2seq models similarly obtain impressive perplexities for word-generation tasks.

Notably, however, seq2seq models are not used as conditional language
models at test-time; they must instead generate fully-formed word sequences. In practice, generation is accomplished by searching over output sequences greedily or with beam search. In this context, \newcite{ranzato16sequence}
note that the combination of the training and generation scheme just described leads to at least two major issues:

\begin{enumerate}
\item \textit{Exposure Bias}: the model is never exposed to its own
  errors during training, and so the inferred histories at test-time
  do not resemble the gold training histories.
\item \textit{Loss-Evaluation Mismatch}: training uses a word-level
  loss, while at test-time we target improving sequence-level
  evaluation metrics, such as BLEU~\cite{papineni02bleu}.
\end{enumerate}

We might additionally add the concern of \textit{label bias}~\cite{lafferty01conditional}
to the list, since word-probabilities at each time-step are locally
normalized, guaranteeing that successors of incorrect histories receive
the same mass as do the successors of the true history.

In this work we develop a non-probabilistic variant of the seq2seq
model that can assign a score to any possible target
\textit{sequence}, and we propose a training procedure, inspired by the learning
as search optimization (LaSO) framework of \newcite{daume05learning},
that defines a loss function in terms of errors made during beam
search. Furthermore, we provide an efficient algorithm to
back-propagate through the beam-search procedure during seq2seq
training.

This approach offers a possible solution to each of the three
aforementioned issues, while largely maintaining the model architecture and
training efficiency of standard seq2seq learning. Moreover, by scoring sequences rather than words, our approach also allows for enforcing hard-constraints on sequence generation \textit{at training time}. To test out the effectiveness of the proposed approach, we develop a general-purpose seq2seq system with beam
search optimization. We run experiments on three very different
problems: word ordering, syntactic parsing, and machine
translation, and compare to a highly-tuned seq2seq system with
attention~\cite{luong15effective}. The version with beam search optimization shows
significant improvements on all three tasks, and particular
improvements on tasks that require difficult search.

\section{Related Work}
\label{sec:relatedwork}
%Similar issues have been raised and addressed by many authors in
%numerous contexts~\cite{}, however the use of seq2seq poses 
%new training challenges that require specific attention. Recently
%there has been some work (scheduled sampling, ballesteros, ranzato,
%google losers)...
The issues of exposure bias and label bias have received much
attention from authors in the structured prediction community, and we
briefly review some of this work here. One prominent approach to
combating exposure bias is that of SEARN~\cite{daume09search}, a
meta-training algorithm that learns a search policy in the form of a
cost-sensitive classifier trained on examples generated from an
interpolation of an oracle policy and the model's current (learned)
policy. Thus, SEARN explicitly targets the mismatch between oracular
training and non-oracular (often greedy) test-time inference by
training on the output of the model's own
policy. DAgger~\cite{ross11a} is a similar approach, which differs in
terms of how training examples are generated and aggregated, and there
have additionally been important refinements to this style of training
over the past several years~\cite{chang15efficient}. When it comes to
training RNNs, SEARN/DAgger has been applied under the name
``scheduled sampling''~\cite{bengio15scheduled}, which involves
training an RNN to generate the $t\,{+}\,1$'st token in a target
sequence after consuming either the true $t$'th token, or, with
probability that increases throughout training, the predicted $t$'th
token.

Though technically possible, it is uncommon to use beam search when
training with SEARN/DAgger. The
early-update~\cite{collins04incremental} and
LaSO~\cite{daume05learning} training strategies, however, explicitly
account for beam search, and describe strategies for updating
parameters when the gold structure becomes unreachable during
search. Early update and LaSO differ primarily in that the former
discards a training example after the first search error, whereas LaSO
resumes searching after an error from a state that includes the gold
partial structure. In the context of feed-forward neural network
training, early update training has been recently explored in a
feed-forward setting by \newcite{zhou15a} and
\newcite{andor16globally}. Our work differs in that we adopt a
LaSO-like paradigm (with some minor modifications), and apply it to
the training of seq2seq RNNs (rather than feed-forward networks). We
also note that \newcite{watanabe15transition} apply
maximum-violation training~\cite{huang12structured}, which is similar
to early-update, to a parsing model with 
recurrent components, and that \newcite{yazdani15incremental} use
beam-search in training a discriminative, locally normalized
dependency parser with recurrent components.

Recently authors have also proposed alleviating exposure bias using
techniques from reinforcement learning. 
\newcite{ranzato16sequence} follow this approach to train RNN decoders
in a seq2seq model, and they obtain consistent improvements in performance, even over models trained with scheduled sampling. As \newcite{daume05learning} note, LaSO is similar to reinforcement learning, except
it does not require ``exploration'' in the same
way. Such exploration may be unnecessary in supervised text-generation, since we typically know the
gold partial sequences at each time-step. \newcite{shen16mrt} use minimum risk training (approximated by sampling) to address the issues of exposure bias and loss-evaluation mismatch for seq2seq MT, and show impressive performance gains.

Whereas exposure bias results from training in a certain way, label
bias results from properties of the model itself. In particular, label
bias is likely to affect structured models that make sub-structure
predictions using locally-normalized scores. Because the neural and
non-neural literature on this point has recently been reviewed by
\newcite{andor16globally}, we simply note here that RNN
models are typically locally normalized, and we are unaware of any
specifically seq2seq work with RNNs that does \textit{not} use
locally-normalized scores. The model we introduce here, however, is
not locally normalized, and so should not suffer from label bias. We
also note that there are some (non-seq2seq) exceptions to the trend of
locally normalized RNNs, such as the work of \newcite{sak14sequence}
and \newcite{voigtlaender15sequence}, who train LSTMs in the context
of HMMs for speech recognition using sequence-level objectives; their
work does not consider search, however.

%and label bias [what about sequence level losses? maybe just forget...]
%
%Split into linear and non-linear.

\section{Background and Notation}
\label{sec:background}
In the simplest seq2seq scenario, we are given a collection of source-target
sequence pairs and tasked with learning to generate
target sequences from source sequences. For instance, we might view machine translation in this way, where in particular we attempt to generate English sentences from (corresponding) French sentences. Seq2seq models are part of the broader class of ``encoder-decoder'' models~\cite{cho14on}, which first use an encoding model to transform a source object into an encoded representation $\boldx$. Many different sequential
(and non-sequential) encoders have proven to be effective for
different source domains. In this work we are agnostic to the
form of the encoding model, and simply assume an abstract source
representation $\boldx$. %In experiments we utilize an attention-based LSTM encoder \cite{} which has shown to be effective for many tasks \cite{}.

Once the input sequence is encoded, seq2seq models generate a target
sequence using a \textit{decoder}. The decoder is tasked with
generating a target sequence of words from a target vocabulary $\mcV$. In particular, words are generated sequentially by conditioning on the input representation $\boldx$ and on the previously generated words or \textit{history}. We use the notation $\pfx{T}$ to refer to an arbitrary word sequence of length $T$, and the notation $\goldpfx{T}$ to refer to the \textit{gold} (i.e., correct) target word sequence for an input $\boldx$. 

Most seq2seq systems utilize a recurrent neural network (RNN) for the decoder model. Formally, a recurrent neural network is a parameterized non-linear
function $\RNN$ that recursively maps a sequence of vectors to a
sequence of hidden states. Let $\boldm_1, \ldots, \boldm_T$ be a
sequence of $T$ vectors, and let $\boldh_0$ be some initial state
vector. Applying an RNN to any such sequence yields hidden states
$\boldh_t$ at each time-step $t$, as follows:
\begin{align*}
\boldh_t \gets \RNN(\boldm_t, \boldh_{t-1}; \btheta),
\end{align*}
where $\btheta$ is the set of model parameters, which are shared over time. In this work, the vectors $\boldm_t$ will always correspond to the embeddings of a target word sequence $\pfx{T}$, and so we will also write $\boldh_t \gets \RNN(w_t, \boldh_{t-1}; \btheta)$, with $w_t$ standing in for its embedding.
 
%To back-propagate errors through a recurrent neural network, we accumulate the 
%gradients of each state with respect to subsequent states by running a backward procedure we will refer to as $\BRNN$ at each time-step (starting at the penultimate step): 
%\begin{align*}
%\nabla_{\boldh_t} \mcL \gets \BRNN(y_{t+1}, \boldh_{t},\nabla_{\boldh_{t+1}} \mcL),
%\end{align*}
%$\BRNN$ takes into account $\boldh_t$'s contribution to any loss incurred from its next-step prediction, as well as to any loss incurred through $\boldh_{t+1}$. In what follows, we will often abbreviate $\nabla_{\boldh_t} \mcL$ as $\nabla_{\boldh_t}$.  
%%\begin{align*}
%%\nabla_{\boldh_t} \mcL \gets \nabla_{\boldh_t} \mcL + \BRNN(\nabla_{\boldh_{t+1}} \mcL, \boldm_t, \boldh_{t}).
%%\end{align*}
%%Note that $\boldm_t$ is the embedding corresponding to output word $w_t$. 
%Running this $\BRNN$ procedure from $t \niceq T$ to $t \niceq 1$ is known as back-propagation through time (BPTT).

%\textbf{something about BPTT}

%  which takes the form of a recurrent
% neural network (RNN). 

% where a
% decoder RNN generates a target sequence of T
% words w1 · · · wT (such as a translation or summary),
% from an

% As RNN decoding is the main focus of this work,
% we now describe this process in greater detail.  

RNN decoders are typically trained to act as conditional language
models. That is, one attempts to model the probability of the $t$'th target
word conditioned on $\boldx$ and the target history by stipulating that $p(w_{t} | \pfx{t-1}, \boldx) \niceq g(w_{t},
\boldh_{t-1}, \boldx)$, for some parameterized function $g$ typically computed with an affine layer followed by a softmax. In computing these probabilities, the state $\boldh_{t-1}$ represents the target history, and $\boldh_0$ is typically set to be some function of $\boldx$. The complete model (including encoder) is trained,
analogously to a neural language model, to minimize the cross-entropy
loss at each time-step while conditioning on the gold history in the
training data. That is, the model is trained to minimize $-\ln \prod_{t=1}^{T} p(y_{t} |\goldpfx{t-1}, \boldx)$.

Once the decoder is trained, discrete sequence generation can be
performed by approximately maximizing the probability of the target
sequence under the conditional distribution,
$\hat{y}_{1:T} \niceq \mathrm{argbeam}_{w_{1:T}} \prod_{t=1}^{T} p(w_t |\pfx{t-1}, \boldx)$, where we use the notation $\mathrm{argbeam}$ to emphasize that the decoding process requires heuristic search, since the RNN model is non-Markovian. In practice, a simple beam search
procedure that explores $K$ prospective histories at each time-step
has proven to be an effective decoding approach. However, as noted above,
decoding in this manner after conditional language-model style training \textit{potentially} suffers from the issues of exposure bias and label bias, which motivates the work of this paper.

\section{Beam Search Optimization}
We begin by making one small change to the seq2seq modeling
framework. Instead of predicting the probability of the next
word, we instead learn to produce (non-probabilistic) scores
for ranking sequences. Define the score of a sequence consisting of 
\textit{history} $\pfx{t-1}$ followed by a single word $w_{t}$ as $f(w_{t}, \boldh_{t-1}, \boldx)$,
%\begin{align} \label{eq:score}
%\score(\pfx{t}, w_{t+1}) \triangleq f(w_{t+1}, \boldh_t, \boldx),
%\end{align} 
where $f$ is a parameterized function examining the current
hidden-state of the relevant RNN at time $t\,{-}\,1$ as well as the input
representation $\boldx$. In experiments, our $f$ will have an identical 
form to $g$ but \textit{without} the final softmax transformation (which transforms unnormalized scores into probabilities),  thereby allowing the model to avoid issues associated with the label bias
problem.

% employing a model that assigns
%a score to global sequences as opposed to probabilities to local
%decision avoids issues associated with the label bias
%problem

% Note that we use $\cdot$ as the concatenation
% operator. 

%As discussed in section~\ref{related}, employing a model that assigns
%a score to global sequences as opposed to probabilities to local
%decision avoids issues associated with the label bias
%problem. Unfortunately exactly computing gradients in standard
%sequence-level models requires performing an intractable search
%problem in training. [So his point here is that finding actual top K is infeasible; we can make the point that non-probabilistic loss allows us to not approximate] As seq2seq training is computationally very
%demanding to start, any major slowdown makes the training process
%infeasible.

More importantly, we also modify how this model is
trained. Ideally we would train by comparing the gold sequence to the
highest-scoring complete sequence. However, because finding the
argmax sequence according to this model is intractable, we
propose to adopt a LaSO-like~\cite{daume05learning} scheme to train, which we will refer to as beam search optimization (BSO). In particular, we define a loss that
penalizes the gold sequence falling off the beam during
training.\footnote{Using a non-probabilistic model further allows us
  to incur no loss (and thus require no update to parameters) when the
  gold sequence \textit{is} on the beam; this contrasts with models
  based on a CRF loss, such as those of \newcite{andor16globally} and
  \newcite{zhou15a}, though in training those models are simply not
  updated when the gold sequence remains on the beam.} The proposed
training approach is a simple way to expose the model to incorrect
histories and to match the training procedure to test
generation. Furthermore we show that it can be implemented efficiently
without changing the asymptotic run-time of training, beyond a factor
of the beam size $K$.

%Instead of directly optimizing our true loss,
%LaSO acknowledges that we will not be able to find the optimal target
%sequence even at test time. Therefore we instead train the model to
%correctly learn to rank sequences within beam search during training.

% using this model also requires a global training procedure. 

% of arbitrary \textit{sequences} formed from
% the target vocabulary $\mcV$. Weg will accordingly 

\subsection{Search-Based Loss}
We now formalize this notion of a search-based loss for RNN training. Assume we have a
set $S_t$ of $K$ candidate sequences of length $t$. We can calculate a
score for each sequence in $S_t$ using a scoring function $f$
parameterized with an RNN, as above, and we define the
sequence $\beampred{t}{K} \nicein S_t$ to be the $K$'th
ranked sequence in $S_t$ according to $f$. That is, assuming
distinct scores, 
\begin{align*}
\small
|\{ \beampred{t}{k} \nicein S_t \mid f(\hat{y}_t^{(k)}, \hat{\boldh}_{t-1}^{(k)}) > f(\hat{y}_t^{(K)}, \hat{\boldh}_{t-1}^{(K)})\}|
=
K\,{-}\,1,
\end{align*}
where $\hat{y}_t^{(k)}$ is the $t$'th token in $\beampred{t}{k}$,
$\hat{\boldh}_{t-1}^{(k)}$ is the RNN state corresponding to its
$t\,{-}\,1$'st step, and where we have omitted the $\boldx$ argument to $f$ for brevity. %$\beampred{t+1}{k}$ has a higher score than all but $k-1$ other sequences in $S_t$.
%That is (assuming all sequences in $\suk(\pfx{t})$ have distinct scores), we have 
%\begin{align*}
%|\{s \in \suk(\pfx{t}) \mid \score(s) > \score(\hat{\boldy}_{1:t+1}^{(k)})) \}| = k-1
%\end{align*}
%\mcL(f) =

We now define a loss function that gives loss each time the score of
the gold prefix $\goldpfx{t}$ does not exceed that of
$\beampred{t}{K}$ by a margin:
\begin{align*}
 \mcL&(f) = \\
 &\sum_{t=1}^T \Delta(\beampred{t}{K}) \left[1 - f(y_t, \boldh_{t-1}) + f(\hat{y}_t^{(K)},\hat{\boldh}_{t-1}^{(K)}) \right] .
\end{align*}
Above, the $\Delta(\beampred{t}{K})$ term denotes a
mistake-specific cost-function, which allows us to scale the loss
depending on the severity of erroneously predicting $\beampred{t}{K}$;
it is assumed to return 0 when the margin requirement is satisfied,
and a positive number otherwise. It is this term that allows us to use sequence- rather than word-level costs in training (addressing the 2nd issue in the introduction). For instance, when training a seq2seq model for machine translation, it may be desirable to have $\Delta(\beampred{t}{K})$ be inversely related to the partial sentence-level BLEU score of $\beampred{t}{K}$ with $\goldpfx{t}$; we experiment along these lines in Section~\ref{sec:tasks}. 

Finally, because we want the full gold sequence to be at the top of the beam at the end of search, when $t \niceq T$ we modify the loss to require the score of $\goldpfx{T}$ to exceed the score of the \textit{highest} ranked incorrect prediction by a margin.

We can optimize the loss $\mcL$ using a two-step process: (1) in a forward pass, we compute candidate sets $S_t$ and record margin violations (sequences with non-zero loss); (2) in a backward pass, we back-propagate the errors through the seq2seq RNNs.
%\begin{itemize}
%\item compute candidate sets $S_t$ and collect losses 
%\item back-propagate the errors through the seq2seq RNNs and update the parameters 
%\end{itemize}
Unlike standard seq2seq training, the first-step requires running
search (in our case beam search) to find margin violations. The second
step can be done by adapting back-propagation through time (BPTT). 
We next discuss the details of this process.

% The first-step requires the 

% \textbf{[When is the right time to note that at the final step we require gold to actually be first on beam?]}

\subsection{Forward: Find Violations} 
\label{sec:forward}
In order to minimize this loss, we need to specify a procedure for
constructing candidate sequences $\beampred{t}{k}$ at each time step
$t$ so that we find margin violations. We follow LaSO (rather than
early-update \footnote{We found that training with early-update rather than (delayed)
LaSO did not work well, even after pre-training. Given the success of early-update
in many NLP tasks this was somewhat surprising. We leave this question to future work.}; see Section~\ref{sec:relatedwork}) and build candidates
in a recursive manner. If there was no margin violation at $t{-}1$,
then $S_t$ is constructed using a standard beam search update. If
there was a margin violation, $S_t$ is constructed as the $K$ best
sequences assuming the gold history $y_{1:t-1}$ through time-step $t{-}1$.

%  that violate this
% margin criteria. To do this calculate $\beampred{t}{K}$ at each
% time-step $t$ with respect to the set $S_t$ consisting of candidate
% sequences of length $t$ that can be formed from the $K$ highest
% scoring sequences generated at the $t\,{-}\,1$'st step of beam-search
% \textit{if we incurred no loss at time $t\,{-}\,1$}; otherwise $S_t$
% consists of only those candidates that can be formed from
% $\goldpfx{t}$. 

Formally, assume the
function $\suk$ maps a sequence $\pfx{t-1} \nicein \mcV^{t-1}$ to the
set of all valid sequences of length $t$ that can be formed by appending to it a
valid word $w \nicein \mcV$. In the
simplest, unconstrained case, we will have
\begin{align*}
\suk(\pfx{t-1}) = \{\pfx{t-1}, w \mid w \in \mcV \}.
\end{align*}

As an important aside, note that for some problems it may be preferable to define a
$\suk$ function which imposes hard constraints on successor sequences. For instance, if we would like to use seq2seq models for parsing (by emitting a constituency or dependency structure encoded into a sequence in some way), we will have hard constraints on the sequences the model can output, namely, that they represent valid parses. While hard constraints such as these would be difficult to add to standard seq2seq at training time, in our framework they can naturally be added to the $\suk$ function, allowing us to \textit{train} with hard constraints; we experiment along these lines in Section~\ref{sec:tasks}, where we refer to a model trained with constrained beam search as ConBSO. 

\begin{figure}[t!]
\centering
\includegraphics[width=1\columnwidth]{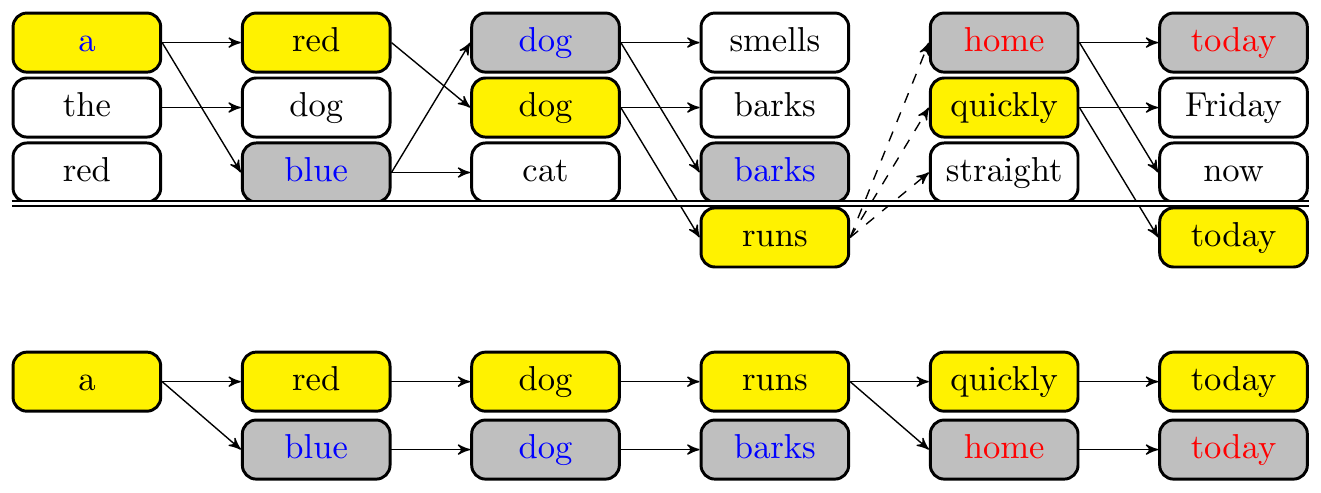}
\caption{Top: possible $\beampred{t}{k}$ formed in training with a beam of size $K \niceq 3$ and with gold sequence $\goldpfx{6}$ = ``a red dog runs quickly today''. The gold sequence is highlighted in yellow, and the predicted prefixes involved in margin violations (at $t \niceq 4$ and $t \niceq 6$) are in gray. Note that time-step $T \niceq 6$ uses a different loss criterion. Bottom: prefixes that actually participate in the loss, arranged to illustrate the back-propagation process.}
\label{fig:backprop}
\end{figure}

Having defined an appropriate $\suk$ function, we specify the candidate set as: 
\begin{align*} \label{eq:stupdate}
S_t =  \topK\begin{cases}  \suk(\goldpfx{t-1}) &\mbox{violation at } t{-}1\\
\bigcup_{k=1}^K \suk(\beampred{t-1}{k}) &\mbox{otherwise}, \end{cases}
\end{align*}
where we have a margin violation at $t{-}1$ iff $f(y_{t-1}, \boldh_{t-2}) < f(\hat{y}_{t-1}^{(K)},\hat{\boldh}_{t-2}^{(K)}) + 1$, and where $\topK$ considers the scores given by $f$. This search procedure is illustrated in the top portion of Figure~\ref{fig:backprop}. 

In the forward pass of our training algorithm, shown as the first part of Algorithm~\ref{alg:treebp}, we run this version of
beam search and collect all sequences and their hidden states that
lead to losses. 

\subsection{Backward: Merge Sequences}
Once we have collected margin violations we can run backpropagation to compute parameter updates. Assume a
margin violation occurs at time-step $t$ between the predicted history
$\beampred{t}{K}$ and the gold history $\goldpfx{t}$. As in standard
seq2seq training we must back-propagate this error through the gold
history; however, unlike seq2seq we also have a gradient for the
wrongly predicted history.

Recall that to back-propagate errors through an RNN we run a recursive backward procedure --- denoted below by $\BRNN$ --- at each time-step $t$, which accumulates the 
gradients of next-step and future losses with respect to $\boldh_t$. We have: 
\begin{align*}
\nabla_{\boldh_t} \mcL \gets \BRNN(\nabla_{\boldh_t} \mcL_{t+1},\nabla_{\boldh_{t+1}} \mcL),
\end{align*} 
where $\mcL_{t+1}$ is the loss at step $t \, {+} \, 1$, deriving, for instance, from the score $f(y_{t+1}, \boldh_{t})$.
%\begin{align*}
%\nabla_{\boldh_t} \mcL \gets \nabla_{\boldh_t} \mcL + \BRNN(\nabla_{\boldh_{t+1}} \mcL, \boldm_t, \boldh_{t}).
%\end{align*}
%Note that $\boldm_t$ is the embedding corresponding to output word $w_t$. 
Running this $\BRNN$ procedure from $t \niceq T \, {-} \,1$ to $t \niceq 0$ is known as back-propagation through time (BPTT).

In determining the total computational cost of back-propagation here, first note that in the worst case there is one violation at each time-step, which
leads to $T$ independent, incorrect sequences. Since we need to call $\BRNN$
$O(T)$ times for each sequence, a naive strategy of running BPTT for each incorrect sequence would lead to an $O(T^2)$ backward pass, rather than the $O(T)$ time required for the standard seq2seq approach. 

Fortunately, our combination of search-strategy and loss make it
possible to efficiently share $\BRNN$ operations. This shared
structure comes naturally from the LaSO update, which resets the beam in a convenient way.

% Note though that this differs from standard 
% seq2seq training in that we have gradients for   

% Using this search procedure

% An online implementation of the training scheme implied by the loss
% and search-strategy discussed above would then involve beam-searching
% in linear order for the target sequence, using \eqref{eq:stupdate} to
% determine the set of candidates $S_t$ at each time-step. If a margin
% violation occurs at time-step $t$, errors would then be backpropagated
% through both the predicted prefix $\beampred{t}{K}$ and the gold
% prefix $\goldpfx{t}$. The problem with such an approach, however, is
% that it has a worst-case quadratic dependence on $T$, the length of
% the target sequence, since it backpropagates for every violating
% prefix. 

We informally illustrate the process in Figure~\ref{fig:backprop}. The
top of the diagram shows a possible sequence of $\beampred{t}{k}$
formed during search with a beam of size 3 for the target sequence
$y=$ ``a red dog runs quickly today.'' When the gold sequence falls
off the beam at $t \niceq 4$, search resumes with
$S_5 \niceq \suk(\goldpfx{4})$, and so all subsequent predicted
sequences have $\goldpfx{4}$ as a prefix and are thus functions of
$\boldh_4$. Moreover, because our loss function only involves the
scores of the gold prefix and the violating prefix, we end up with the
relatively simple computation tree shown at the bottom of
Figure~\ref{fig:backprop}. It is evident that we can backpropagate
in a single pass, accumulating gradients from sequences that diverge
from the gold at the time-step that precedes their divergence. The
second half of Algorithm~\ref{alg:treebp} shows this explicitly for a
single sequence, though it is straightforward to extend the algorithm
to operate in batch.\footnote{We also note that because
  we do not update the parameters until after the $T$'th search step, our training procedure
  differs slightly from LaSO (which is online), and in this aspect is
  essentially equivalent to the ``delayed LaSO update'' of \newcite{BandK:14}.}
{\small
\begin{algorithm}[t!]
  \small
  \begin{algorithmic}[1]
    \Procedure{BSO}{$\boldx, K_{tr}, \suk$}
    \State{/*\textsc{Forward}*/}
    \State{Init empty storage $\hat{y}_{1:T}$ and $\hat{\boldh}_{1:T}$; init $S_1$}
    \State{$r \gets 0$; $violations \gets \{0\}$}
    \For{$t=1,\ldots,T$}
    \State{$K \niceq K_{tr}$ if $t \, {\neq} \,T$ else $\displaystyle \argmax_{k: \beampred{t}{k} \neq \goldpfx{t}} {\footnotesize f(\hat{y}_t^{(k)}, \hat{\boldh}_{t-1}^{(k)})}$}
    \If{$f(y_{t}, \boldh_{t-1}) < f(\hat{y}_t^{(K)},\hat{\boldh}_{t-1}^{(K)}) + 1$}  
    \State{$\hat{\boldh}_{r:t-1} \gets \hat{\boldh}^{(K)}_{r:t-1}$}
    \State{$\hat{y}_{r+1:t} \gets \hat{y}^{(K)}_{r+1:t}$}    
    \State{Add $t$ to $violations$}
    \State{$r \gets t$}
    \State{$S_{t+1} \gets \topK(  \suk(\goldpfx{t}))$}
    \Else{}
    \State{$S_{t+1} \gets \topK( \bigcup_{k=1}^K \suk(\beampred{t}{k})) $}
    \EndIf{}
    
    \EndFor{}
    \State{/*\textsc{Backward}*/}
%    \State{$\nabla_{\boldh_T} \gets - \Delta(\beampred{T}{K}) \times \nabla_{\boldh_T} f(\goldpfx{T})$}    
%    \State{$\nabla_{\widehat{\boldh}_T} \gets \Delta(\beampred{T}{K}) \times \nabla_{\widehat{\boldh}_T} f(\beampred{T}{K})$}    
%    \State{$\nabla_{\boldh_{T}} \gets \bzero$; $\nabla_{\widehat{\boldh}_{T}} \gets \bzero$}
    \State{$grad\_{\boldh_{T}} \gets \bzero$; $grad\_{\widehat{\boldh}_{T}} \gets \bzero$}
    \For{$t=T-1,\ldots,1$}
    \State{$grad\_{\boldh_t} \, {\gets} \, \BRNN(\nabla_{\boldh_t} \mcL_{t+1}, grad\_{\boldh_{t+1}})$
      }
    \State{$grad\_{\widehat{\boldh}_t} \, {\gets} \, 
    \BRNN(\nabla_{\widehat{\boldh}_t} \mcL_{t+1}, grad\_{\widehat{\boldh}_{t+1}})$
      }      

    \If{$t \, {-} \, 1 \in violations$}
    \State{$grad\_{\boldh_t} \gets grad\_{\boldh_t} + grad\_{\widehat{\boldh}_t}$}     
    \State{$grad\_{\widehat{\boldh}_t} \gets \bzero$ }
    \EndIf{}
    \EndFor{}
    %\State{Update RNN params based on $\boldh, \hat{\boldh}$ }
    \EndProcedure{}
  \end{algorithmic}
  \caption{\label{alg:treebp} Seq2seq Beam-Search Optimization}
\end{algorithm}
}

\section{Data and Methods}
We run experiments on three different tasks, comparing our approach to
the seq2seq baseline, and to other relevant baselines.

\subsection{Model}
While the method we describe applies to seq2seq RNNs in general,
for all experiments we use the global attention model of \newcite{luong15effective} ---
which consists of an LSTM~\cite{hochreiter1997lstm} encoder and an
LSTM decoder with a global attention model --- as both the baseline
seq2seq model (i.e., as the model that computes the $g$ in
Section~\ref{sec:background}) and as the model that computes our
sequence-scores $f(w_{t}, \boldh_{t-1}, \boldx)$. As in \newcite{luong15effective}, we also use
``input feeding,'' which involves feeding the attention distribution
from the previous time-step into the decoder at the current step. 
This model architecture has been found to be highly performant for neural
machine translation and other seq2seq tasks. 

To distinguish the models we refer to our system as BSO (beam search
optimization) and to the baseline as seq2seq. When we apply constrained training (as discussed in Section~\ref{sec:forward}), we refer to the model as ConBSO. In providing results we also distinguish between the
beam size $K_{tr}$ with which the model is trained, and the beam size
$K_{te}$ which is used at test-time. In general, if we plan on
evaluating with a beam of size $K_{te}$ it makes sense to train with a
beam of size $K_{tr} = K_{te} \, {+} \, 1$, since our objective requires the
gold sequence to be scored higher than the \textit{last} sequence on
the beam.

\subsection{Methodology}
Here we detail additional techniques we found necessary to ensure the model learned effectively. First, we found that the model failed to learn when trained
from a random initialization.\footnote{This may be because there is relatively little signal in the sparse, sequence-level gradient, but this point requires further investigation.} We therefore found it necessary to pre-train the model using a standard,
word-level cross-entropy loss as described in
Section~\ref{sec:background}. The necessity of pre-training in this instance is consistent with the findings of other authors who train non-local neural models~\cite{kingsbury09lattice,sak14sequence,andor16globally,ranzato16sequence}.\footnote{\newcite{andor16globally} found, however, that pre-training only increased convergence-speed, but was not necessary for obtaining good results.}

Similarly, it is clear that the smaller the beam used in training is, the less room the model has to make erroneous predictions without running afoul of the margin loss. Accordingly, we also found it useful to use a ``curriculum beam'' strategy in training, whereby the size of the beam  is increased gradually during training. In particular, given a desired training beam size $K_{tr}$, we began training with a beam of size 2, and increased it by 1 every 2 epochs until reaching $K_{tr}$. 

Finally, it has been established that \textit{dropout}~\cite{srivastava14dropout} regularization improves the performance of LSTMs~\cite{pham14dropout,zaremba14rnn}, and in our experiments we run beam search under dropout.\footnote{However, it is important to ensure that the same mask applied at each time-step of the forward search is also applied at the corresponding step of the backward pass. We accomplish this by pre-computing masks for each time-step, and sharing them between the partial sequence LSTMs.}

For all experiments, we trained both seq2seq and BSO  models with mini-batch Adagrad~\cite{duchi2011adaptive} (using batches of size 64), and we renormalized all gradients so they did not exceed 5 before updating parameters. We did not extensively tune learning-rates, but we found initial rates of 0.02 for the encoder and decoder LSTMs, and a rate of 0.1 or 0.2 for
the final linear layer (i.e., the layer tasked with making word-predictions at each
time-step) to work well across all the tasks we considered. Code implementing the experiments described below can be found at \url{https://github.com/harvardnlp/BSO}.\footnote{Our code is based on Yoon Kim's seq2seq code, \url{https://github.com/harvardnlp/seq2seq-attn}.}

\subsection{Tasks and Results}
\label{sec:tasks}
Our experiments are primarily intended to evaluate the
effectiveness of beam search optimization over standard seq2seq training. As such, we run experiments with the same model across three very different problems: word ordering, dependency parsing, and machine translation. While we do not include all the features and extensions necessary to
reach state-of-the-art performance, even the baseline seq2seq model is generally quite performant.

\paragraph{Word Ordering}
The task of correctly ordering the words in a shuffled sentence has recently gained some
attention as a way to test the (syntactic) capabilities of
text-generation
systems~\cite{zhang11syntax,zhang15discriminative,liu15transition,schmaltz16word}. 
We cast this task as seq2seq problem by viewing a shuffled sentence as
a source sentence, and the correctly ordered sentence as the
target. While word ordering is a somewhat synthetic task, it has two interesting properties for our purposes. First, it is a task which plausibly requires search (due to the exponentially many possible orderings), and, second, there is a clear hard constraint on output sequences, namely, that they be a permutation of the source sequence. For both the baseline and BSO models we
enforce this constraint at test-time. However, we also experiment with
constraining the BSO model during training, as described in Section~\ref{sec:forward}, by defining the $\suk$ function to only allow successor sequences containing un-used words in the source sentence.

For experiments, we use the same PTB dataset (with the standard training,
development, and test splits) and evaluation procedure as in
\newcite{zhang15discriminative} and later work, with performance reported in terms of BLEU
score with the correctly ordered sentences. For all word-ordering
experiments we use 2-layer encoder and decoder LSTMs, each with 256
hidden units, and dropout with a rate of 0.2 between LSTM layers. We use simple 0/1 costs in defining the $\Delta$ function.  

We show our test-set results
in Table~\ref{tab:wo}. We see that on this task there is a large improvement at 
each beam size from switching to BSO, and a further improvement from using 
the constrained model.

\begin{table}
  \centering
  \begin{tabular}{lccc}
    \toprule
     & \multicolumn{3}{c}{Word Ordering (BLEU) } \\ 
          & $K_{te}$ = 1 & $K_{te}$ = 5 & $K_{te}$ = 10 \\ 
    \midrule
    seq2seq & 25.2 & 29.8 & 31.0 \\
    BSO     & 28.0 & 33.2 & 34.3 \\
    ConBSO & \textbf{28.6} & \textbf{34.3} & \textbf{34.5} \\
    \midrule
    LSTM-LM & 15.4 &  - & 26.8 \\
    \bottomrule
  \end{tabular}
  \caption{Word ordering. BLEU Scores of seq2seq, BSO, constrained BSO, and a vanilla LSTM language model (from Schmaltz et al, 2016). All experiments above have $K_{tr}\,{=}\,6$.}
  \label{tab:wo}
\end{table}

Inspired by a similar analysis in \newcite{daume05learning}, we further examine the relationship between $K_{tr}$ and $K_{te}$ when training with ConBSO in Table~\ref{tab:wosizeexp}. We see that larger $K_{tr}$ hurt greedy inference, but that results continue to improve, at least initially, when using a $K_{te}$ that is (somewhat) bigger than $K_{tr}-1$.
\begin{table}
  \centering
  \begin{tabular}{lccc}
    \toprule
    & \multicolumn{3}{c}{Word Ordering Beam Size (BLEU) } \\ 
    &  $K_{te}$ = 1 & $K_{te}$ = 5 & $K_{te}$ = 10 \\ 
    \midrule
    $K_{tr}$ = 2 & 30.59 & 31.23 & 30.26 \\
    $K_{tr}$ = 6 & 28.20 & 34.22 & 34.67 \\
    $K_{tr}$ = 11 & 26.88 & 34.42 & 34.88 \\   
    \midrule
    seq2seq & 26.11 & 30.20 & 31.04 \\         
    \bottomrule
  \end{tabular}
  \caption{Beam-size experiments on word ordering development set. All numbers reflect training with constraints (ConBSO).}
  \label{tab:wosizeexp}
\end{table}
% There, we compare with....

\paragraph{Dependency Parsing}
% (in the style of \newcite{chen14fast} and \newcite{weiss15structured})
We next apply our model to dependency parsing, which also has hard constraints and plausibly benefits from search. We treat dependency parsing with arc-standard transitions as a seq2seq task by attempting to map from a source sentence to a target sequence of source sentence words interleaved with the arc-standard, reduce-actions in its parse. For example, we attempt to map the source sentence \begin{quote}
But it was the Quotron problems that  ...
\end{quote} to the target sequence \begin{quote}
But it was @L\_SBJ @L\_DEP the Quotron problems @L\_NMOD @L\_NMOD that ...
\end{quote}
We use the standard Penn Treebank dataset splits with Stanford dependency labels, and the standard UAS/LAS evaluation metric (excluding punctuation) following  \newcite{chen14fast}. All models thus see only the words in the source and, when decoding, the actions it has emitted so far; no other features are used. We use 2-layer encoder and decoder LSTMs with 300 hidden units per layer and dropout with a rate of 0.3 between LSTM layers. We replace singleton words in the training set with an UNK token, normalize digits to a single symbol, and initialize word embeddings for both source and target words from the publicly available \texttt{word2vec}~\cite{mikolov2013distributed} embeddings. We use simple 0/1 costs in defining the $\Delta$ function. 

As in the word-ordering case, we also experiment with modifying the $\suk$ function in order to train under hard constraints, namely, that the emitted target sequence be a valid parse. In particular, we constrain the output at each time-step to obey the stack constraint, and we ensure words in the source are emitted in order. 

We show results on the test-set in Table~\ref{tab:dep}. BSO and ConBSO both show significant improvements over seq2seq, with ConBSO improving most on UAS, and BSO improving most on LAS. We achieve a reasonable final score of 91.57 UAS, which lags behind the state-of-the-art, but is promising for a general-purpose, word-only model.
\begin{table}
  \centering
  %\hspace*{-0.3cm}
  \begin{tabular}{@{}l@{\hspace{4pt}}ccc}
    \toprule
    & \multicolumn{3}{c}{Dependency Parsing (UAS/LAS) } \\ 
          & $K_{te}$ = 1 & $K_{te}$ = 5 & $K_{te}$ = 10 \\ 
    \midrule
    seq2seq & \textbf{87.33/82.26} & 88.53/84.16 & 88.66/84.33\\
    BSO & 86.91/82.11 & 91.00/\textbf{87.18} & 91.17/\textbf{87.41} \\
    ConBSO & 85.11/79.32 & \textbf{91.25}/86.92 & \textbf{91.57}/87.26 \\
    \midrule
    Andor & 93.17/91.18 & - & - \\ 
    \bottomrule
  \end{tabular}
  \caption{Dependency parsing. UAS/LAS of  seq2seq, BSO, ConBSO and baselines on PTB test set. Andor is the current state-of-the-art model for this data set (Andor et al. 2016), and we note that with a beam of size 32 they obtain 94.41/92.55. All experiments above have $K_{tr}\,{=}\,6$.}
  \label{tab:dep}
\end{table}

\paragraph{Translation}
We finally evaluate our model on a small machine translation dataset, which allows us to experiment with a cost function that is not 0/1, and to consider other baselines that attempt to mitigate exposure bias in the seq2seq setting. We use the dataset from the work of \newcite{ranzato16sequence}, which uses data from the German-to-English portion of the IWSLT 2014 machine translation evaluation campaign~\cite{cettolo14report}. The data comes from translated TED talks, and the dataset contains roughly 153K training sentences, 7K development sentences, and 7K test sentences. We use the same preprocessing and dataset splits as \newcite{ranzato16sequence}, and like them we also use a single-layer LSTM decoder with 256 units. We also use dropout with a rate of 0.2 between each LSTM layer. We emphasize, however, that while our decoder LSTM is of the same size as that of \newcite{ranzato16sequence}, our results are not directly comparable, because we use an LSTM encoder (rather than a convolutional encoder as they do), a slightly different attention mechanism, and input feeding~\cite{luong15effective}.

For our main MT results, we set $\Delta(\beampred{t}{k})$ to $1 \,{-}\,\mathrm{SB}(\hat{y}_{r+1:t}^{({K})}, y_{r+1:t})$, where $r$ is the last margin violation and $\mathrm{SB}$ denotes smoothed, sentence-level BLEU \cite{chen14systematic}. This setting of $\Delta$ should act to penalize erroneous predictions with a relatively low sentence-level BLEU score more than those with a relatively high sentence-level BLEU score. In Table~\ref{tab:mtfinal} we show our final results and those from \newcite{ranzato16sequence}.\footnote{Some results from personal communication.} While we start with an improved baseline, we see similarly large increases in accuracy as those obtained by DAD and MIXER, in particular when $K_{te} > 1$. 

\begin{table}[t!]
  \centering
  \begin{tabular}{lccc}
    \toprule
    & \multicolumn{3}{c}{Machine Translation (BLEU) } \\ 
    &  $K_{te}$ = 1 & $K_{te}$ = 5 & $K_{te}$ = 10 \\ 
    \midrule
    seq2seq & 22.53 & 24.03 & 23.87 \\
    BSO, SB-$\Delta$ & \textbf{23.83} & \textbf{26.36} & \textbf{25.48} \\
    \midrule
    XENT & 17.74 & 20.10 & 20.28 \\
    DAD & 20.12 & 22.25 & 22.40 \\ 
    MIXER & 20.73 & 21.81 & 21.83 \\    
    \bottomrule
  \end{tabular}
  \caption{Machine translation experiments on test set; results below middle line are from MIXER model of Ranzato et al. (2016). SB-$\Delta$ indicates sentence BLEU costs are used in defining $\Delta$.  XENT is similar to our seq2seq model but with a convolutional encoder and simpler attention. DAD trains seq2seq with scheduled sampling (Bengio et al., 2015). BSO, SB-$\Delta$ experiments above have $K_{tr} \niceq 6$.}
  \label{tab:mtfinal}
\end{table}

We further investigate the utility of these sequence-level costs in Table~\ref{tab:mtdelt}, which compares using sentence-level BLEU costs in defining $\Delta$ with using 0/1 costs.
\begin{table}[t!]
  \centering
  \begin{tabular}{lccc}
    \toprule
    & \multicolumn{3}{c}{Machine Translation (BLEU)} \\ 
     &  $K_{te}$ = 1 & $K_{te}$ = 5 & $K_{te}$ = 10 \\ 
    \midrule
    0/1-$\Delta$ & 25.73  & 28.21 & 27.43  \\  
        SB-$\Delta$ & 25.99  & 28.45 & 27.58 \\  
    \bottomrule
  \end{tabular}
  \caption{BLEU scores obtained on the machine translation development data when training with $\Delta(\beampred{t}{k}) \niceq 1$ (top) and $\Delta(\beampred{t}{k}) \niceq 1 \,{-}\,\mathrm{SB}(\hat{y}_{r+1:t}^{({K})}, y_{r+1:t})$ (bottom), and $K_{tr}$ = 6. }
\label{tab:mtdelt}
\end{table}
We see that the more sophisticated sequence-level costs have a moderate effect on BLEU score.

%Finally, in Table~\ref{tab:mtsizeexp}
%
%\begin{table}
%  \centering
%  \begin{tabular}{lccc}
%    \toprule
%    & \multicolumn{3}{c}{MT Beam Size (BLEU) } \\ 
%    &  $K_{te}$ = 1 & $K_{te}$ = 5 & $K_{te}$ = 10 \\ 
%    \midrule
%    $K_{tr}$ = 2 & 24.86 & 23.20 & 20.54 \\
%    $K_{tr}$ = 6 & 25.73 & 28.21 & 27.43 \\
%    $K_{tr}$ = 11 & 25.03 & 28.42 & 27.92 \\   
%    \midrule
%    seq2seq & 24.90 & 26.34 & 26.02 \\         
%    \bottomrule
%  \end{tabular}
%  \caption{Beam-size experiments on translation development set. All numbers reflect training with $\Delta(\beampred{t}{k})=1$.}
%  \label{tab:mtsizeexp}
%\end{table}
\paragraph{Timing}
Given Algorithm~\ref{alg:treebp}, we would expect training time to increase linearly with the size of the beam. On the above MT task, our highly tuned seq2seq baseline processes an average of 13,038 tokens/second (including both source and target tokens) on a GTX 970 GPU. For beams of size $K_{tr}$ = 2, 3, 4, 5, and 6, our implementation processes on average 1,985, 1,768, 1,709, 1,521, and 1,458 tokens/second, respectively. Thus, we appear to pay an initial constant factor of $\approx 3.3$ due to the more complicated forward and backward passes, and then training scales with the size of the beam. Because we batch beam predictions on a GPU, however, we find that in practice training time scales sub-linearly with the beam-size. 

\section{Conclusion}
We have introduced a variant of seq2seq and an associated beam search training scheme, which addresses exposure bias as well as label bias, and moreover allows for both training with sequence-level cost functions as well as with hard constraints. Future work will examine scaling this approach to much larger datasets.

\section*{Acknowledgments} We thank Yoon Kim for helpful discussions and for providing the initial seq2seq code on which our implementations are based. We thank Allen Schmaltz for help with the word ordering experiments. We also gratefully acknowledge the support of a Google Research Award.

\nocite{bahdanau16an}

\bibliography{beamtrain}
\bibliographystyle{emnlp2016}

\end{document}